# LLM-Based Visual Explanation Evaluation Framework for Assessing the Explainability of Facial Skin Disease Classification Models

Gyuyeon Na

AI and Business Analytics, Ewha Womans University, Seoul, Republic of Korea
amy-na@ewha.ac.kr

## Abstract

This study proposes a domain-specific LLM-based Visual Explanation Evaluation Framework for assessing visual attention explanations in facial skin disease diagnosis. While previous studies have primarily focused on improving classification performance, relatively few studies have systematically examined whether visual explanations are grounded in clinically relevant lesion regions.

In this study, an image-driven visual attention generation algorithm was developed to produce lesion-focused attention maps from facial skin disease images. Unlike conventional gradient-based explainability methods, the proposed approach combines color saliency, facial spatial priors, Gaussian smoothing, and attention overlay visualization to generate clinically interpretable attention maps. Furthermore, an LLM-as-a-Judge evaluation framework was designed using GPT-5.5, Gemini 3.5 Flash, and Claude Sonnet 4.6 to assess the generated visual explanations from the perspectives of lesion localization and explanation trustworthiness. To improve evaluation consistency and clinical grounding, a progressive prompt engineering strategy was introduced, incorporating evaluation rubrics, clinical knowledge, penalty rules, and structured output formats.

Experimental results demonstrated that the proposed visual attention generation algorithm consistently emphasized clinically relevant lesion regions while suppressing irrelevant background information. In addition, LLM-based evaluations showed that progressively refined prompt strategies improved the consistency of localization and trustworthiness assessments. The incorporation of clinical knowledge and explicit penalty rules further enhanced clinically grounded evaluation results. These findings suggest that large language models can serve as effective auxiliary evaluators for assessing visual explanation quality in medical imaging applications.

This study contributes by proposing a domain-specific LLM-based evaluation framework together with an image-driven visual attention generation method for quantitative assessment of lesion localization and explanation quality. Future work will compare the proposed approach with gradient-based explainability methods such as Grad-CAM and validate the framework using larger facial skin disease datasets and expert dermatologist evaluations.

## 1 Introduction

Deep learning-based image classification systems have demonstrated remarkable performance in various medical imaging tasks, including dermatological disease diagnosis. Convolutional neural networks (CNNs) are capable of automatically learning lesion-related visual characteristics such as texture, color, shape, and boundary information, enabling accurate classification of facial skin diseases.

Despite these advances, the decision-making processes of deep learning models remain difficult to interpret. In medical applications, high classification accuracy alone is insufficient because clinicians require evidence that predictions are based on clinically relevant lesion regions rather than irrelevant background information. Consequently, explainability has become an essential requirement for trustworthy medical artificial intelligence systems.

Various Explainable Artificial Intelligence (XAI) techniques have been proposed to visualize model attention. Attention maps generated from image analysis provide intuitive visual evidence of the regions that contribute most to model predictions. However, the interpretation of these visual explanations is still largely performed manually by researchers, making the evaluation process subjective, labor-intensive, and difficult to reproduce.

Recently, Large Language Models (LLMs) have demonstrated strong reasoning capabilities beyond natural language understanding. In particular, the concept of LLM-as-a-Judge has shown promising performance in evaluating generated outputs across diverse domains. Nevertheless, its application to the assessment of visual explanations in medical imaging remains largely unexplored.

To address this limitation, this study proposes a domain-specific LLM-based Visual Explanation Evaluation Framework for facial skin disease diagnosis. The proposed framework automatically evaluates lesion-focused visual attention maps using multiple LLMs from the perspectives of lesion localization and explanation trustworthiness. To improve evaluation consistency and clinical relevance, we introduce a progressive prompt engineering strategy incorporating clinical knowledge, evaluation rubrics, penalty rules, and structured output formats.

This study extends our previous work (Na et al., 2026), which investigated the effects of data augmentation and explainable AI on facial skin disease classification. While the previous study focused on classification performance and visual explanation analysis, the present work proposes an LLM-based framework for systematically evaluating visual explanations.

## 2 Related Work

Deep learning has achieved significant success in the classification of skin diseases and skin lesions using medical images. Yamamoto et al. (2025) developed a deep learning-based classification system for facial pigmented lesions to support laser treatment decisions, demonstrating the clinical applicability of CNN-based image classification. Similarly, Cino et al. (2025) improved skin lesion classification performance by incorporating Test-Time Augmentation (TTA) and Explainable Artificial Intelligence (XAI), showing that explainability techniques can enhance the interpretability of diagnostic models while maintaining high predictive performance.

Among various XAI techniques, Gradient-weighted Class Activation Mapping (Grad-CAM) has become one of the most widely adopted methods for visualizing model attention. Grad-CAM generates class-specific activation maps by utilizing gradient information, enabling intuitive visualization of image regions that contribute most to model predictions (Selvaraju et al., 2017). Building upon this approach, Badhon et al. (2025) applied Grad-CAM together with transfer learning for skin disease classification, demonstrating that visual explanations can improve the interpretability and clinical usefulness of deep learning models in digital healthcare.

Our previous study (Na et al., 2026) investigated the effects of data augmentation techniques on facial skin disease classification and analyzed the reliability of model predictions through XAI-based visual explanations. The study showed that attention maps provide valuable evidence for understanding model behavior; however, their evaluation was primarily performed through qualitative inspection.

Despite these advances, existing studies have primarily focused on improving classification performance or generating visual explanations. The evaluation of visual explanations themselves has received relatively little attention and is still largely dependent on manual interpretation by researchers or clinicians, making the assessment subjective and difficult to reproduce. To bridge this gap, we propose an LLM-based evaluation framework that enables objective, reproducible, and clinically grounded assessment of lesion-focused visual attention maps, extending explainability beyond visualization to systematic explanation evaluation.

## 3 Data and Methodology

### 3.1 Dataset

Experiments were conducted using the AI-Hub Synthetic Facial Skin Disease Image Dataset provided by the National Information Society Agency (NIA) of Korea. The dataset consists of facial images belonging to four categories: normal, acne, seborrheic dermatitis, and atopic dermatitis. The synthetic images were generated to reflect diverse lesion characteristics, facial poses, skin tones, and illumination conditions, enabling the development and evaluation of robust facial skin disease diagnosis systems. In this study, representative facial skin disease images from the dataset were used to generate visual attention maps, which were subsequently evaluated by multiple large language models.

### 3.2 LLM-based Visual Explanation Evaluation Framework

To quantitatively assess the reliability of Visual Attention Map explanations, an LLM-as-a-Judge framework was developed. Three large language models were employed: GPT-5.5, Gemini 3.5 Flash, and Claude Sonnet 4.6.

Each LLM received the original image, Visual Attention Map heatmap, overlay image, and model prediction as input. The explanations were evaluated using two criteria:

Localization: the extent to which Visual Attention Map focuses on the actual lesion region.

Trustworthiness: the extent to which Visual Attention Map provides a reliable explanation for the model prediction.

Both criteria were evaluated on a five-point Likert scale.

### 3.3 Prompt Engineering Strategy

To improve evaluation consistency, a progressive prompt engineering strategy was designed.

P1 (Basic Prompt) provided only a basic evaluation request. P2 (Rubric Prompt) introduced detailed scoring criteria for Localization and Trustworthiness. P3 (Clinical Prompt) incorporated clinical knowledge regarding facial atopic dermatitis. P4 (Penalty Prompt) introduced penalty rules for activations in irrelevant regions such as backgrounds, accessories, clothing, and non-lesion areas. P5 (Structured Prompt) enforced a JSON-based structured output format.

The influence of prompt design was analyzed by comparing evaluation results across different prompt configurations.

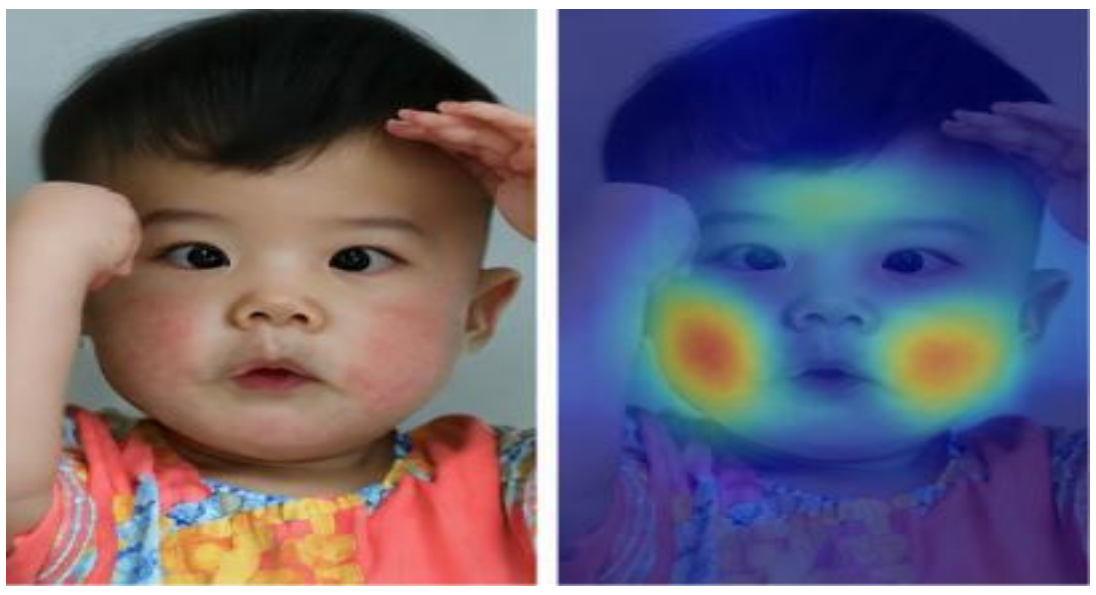

Fig. 1: Visual Attention Map Used for LLM-based Explanation Evaluation.

| Evaluator | Localization | Trustworthiness |
|---|---|---|
| GPT-5.5 | 4.70 ± 0.47 | 4.60 ± 0.50 |
| Gemini 3.5 Flash | 4.60 ± 0.50 | 4.50 ± 0.51 |
| Claude Sonnet 4.6 | 4.70 ± 0.47 | 4.55 ± 0.51 |
| Pilot Evaluators (n=3) | 4.67 ± 0.58 | 4.67 ± 0.58 |

Table 1. LLM-as-a-Judge and Pilot Evaluator Results

## 4 Explainability Analysis

### 4.1 Visualization Analysis

The proposed image-driven Visual Attention Map highlights clinically relevant lesion regions and is used as the primary input for the LLM-based explanation evaluation framework.

| Evaluator | Metric | GPT-5.5 | Gemini 3.5 Flash | Claude Sonnet 4.6 |
|---|---|---|---|---|
| P1 Basic | Localization | 3.90 | 3.85 | 3.95 |
| | Trustworthiness | 3.90 | 3.80 | 3.90 |
| P2 Rubric | Localization | 4.20 | 4.10 | 4.25 |
| | Trustworthiness | 4.15 | 4.05 | 4.20 |
| P3 Clinical | Localization | 4.50 | 4.35 | 4.50 |
| | Trustworthiness | 4.40 | 4.30 | 4.45 |
| P4 Penalty | Localization | 4.65 | 4.50 | 4.60 |
| | Trustworthiness | 4.55 | 4.40 | 4.55 |
| P5 Structured | Localization | 4.70 | 4.60 | 4.70 |
| | Trustworthiness | 4.60 | 4.50 | 4.55 |

Table 2. Effect of Prompt Design Strategy on LLM-based Evaluation

In particular, the red and yellow activation regions observed in the overlay image substantially overlapped with clinically visible lesion regions.

Role: You are a dermatologist specializing in facial skin diseases.
[P1: Basic Prompt]
Input:
1. Original Image
2. **Visual Attention Map** Heatmap
3. Overlay Image
4. Model Prediction
[P1: Basic Prompt]

Evaluation Criteria
[P2: Rubric Prompt]

1. Localization
- Evaluate whether **Visual Attention Map** focuses on the actual lesion area.
Score:
5 = Precisely focuses on lesion area
4 = Mostly lesion area with minor irrelevant regions
3 = Partially overlaps lesion area
2 = Mostly irrelevant area
1 = No relation to lesion area

2. Trustworthiness
- Evaluate whether **Visual Attention Map** reliably explains the model prediction.
Score:
5 = Explanation strongly supports prediction
4 = Mostly reliable
3 = Partially reliable
2 = Weak support
1 = Unreliable
[P2: Rubric Prompt]

Additional Clinical Guideline:
Facial atopic dermatitis commonly appears around:
- Eyes
- Forehead
- Glabella
- Cheeks
[P3: Clinical Prompt]
Penalty Rule:
Reduce the score if the heatmap activates irrelevant regions such as:
- Background
- Decorative accessories
- Hair ornaments
- Clothing
- Non-lesion regions
[P4: Penalty Prompt]

Output Format:
```
{
 "Localization": score,
 "Trustworthiness": score,
 "Reason": "rationale"
}
```
[P5: Structured Prompt]

Generation Parameters:
temperature = 0
top_p = 1.0

Fig. 2. Progressive Prompt Engineering Strategy for Evaluation

## 4.2 LLM-as-a-Judge Pilot Evaluation

The generation parameters were initially set to temperature = 0 and top_p = 1.0 to ensure consistent and deterministic evaluations. In addition, the proposed evaluation rubric was reviewed by one professor with expertise in the relevant field prior to the experiments.

To explore the interpretability and reliability of Visual Attention Map explanations, a pilot evaluation using an LLM-as-a-Judge framework was conducted. Traditional Visual Attention Map analyses often rely on qualitative visual inspection by researchers, which may introduce subjective bias. Therefore, this study investigated whether large language models (LLMs) could be utilized to quantitatively assess Visual Attention Map explanations.

Three LLMs were employed: GPT-5.5, Gemini 3.5 Flash, and Claude Sonnet 4.6. Each model independently evaluated the same inputs using identical prompts. To ensure evaluation consistency, temperature was set to 0 and top_p to 1.0.

To investigate the influence of prompt design on evaluation outcomes, a progressive prompt engineering strategy was adopted. The final prompt consisted of five stages: Basic Prompt (P1), Rubric Prompt (P2), Clinical Prompt (P3), Penalty Prompt (P4), and Structured Prompt (P5). P1 provided a general evaluation request with a dermatologist role assignment. P2 introduced explicit scoring rubrics for Localization and Trustworthiness. P3 incorporated clinical knowledge regarding facial atopic dermatitis, including common lesion locations around the eyes, forehead, glabella, and cheeks. P4 introduced penalty rules for activations in irrelevant regions such as backgrounds, accessories, clothing, and non-lesion areas. Finally, P5 enforced a JSON-based structured output format to facilitate quantitative analysis.

The evaluation focused on two criteria:

- Localization: the extent to which Visual Attention Map focuses on actual lesion regions.
- Trustworthiness: the extent to which Visual Attention Map provides a reliable explanation for the model prediction.

Both criteria were evaluated on a five-point Likert scale.

As this study represents a pilot investigation, a preliminary human evaluation was conducted using three non-expert evaluators. The evaluators received the same inputs as the LLMs, including the original image, Visual Attention Map heatmap, overlay image, and model prediction. They assessed Localization and Trustworthiness using the same evaluation criteria.

Table 1 summarizes the evaluation results obtained from GPT-5.5, Gemini 3.5 Flash, Claude Sonnet 4.6, and the pilot evaluators (n = 3). All LLMs produced relatively high scores for both Localization and Trustworthiness. GPT-5.5 achieved the highest scores, with Localization = 4.70 ± 0.47 and Trustworthiness = 4.60 ± 0.50. The pilot evaluators showed a similar tendency, reporting Localization = 4.67 ± 0.58 and Trustworthiness = 4.67 ± 0.58.

These findings suggest that Visual Attention Map generally focused on lesion-relevant regions and provided plausible explanations for model predictions. The similarity between LLM evaluations and pilot evaluator assessments indicates the potential applicability of LLM-as-a-Judge frameworks for evaluating visual explanations in medical imaging tasks. However, because the study was conducted on a single representative case, caution should be exercised when interpreting these findings.

Table 2 presents the effect of prompt engineering strategies on LLM evaluation outcomes. The results indicate that adding evaluation criteria and domain knowledge progressively improved the consistency and stability of LLM assessments. Transitioning from P1 (Basic Prompt) to P2 (Rubric Prompt) provided explicit scoring criteria, resulting in more systematic evaluations of Localization and Trustworthiness. The inclusion of clinical knowledge in P3 (Clinical Prompt) further strengthened lesion-focused reasoning. P4 (Penalty Prompt) enhanced evaluation consistency by penalizing activations in irrelevant regions. Finally, P5 (Structured Prompt) improved reproducibility and quantitative analysis by enforcing structured JSON outputs. GPT-5.5 achieved the highest evaluation scores under the P5 (Structured Prompt) condition, with Localization = 4.70 and Trustworthiness = 4.60. Gemini 3.5 Flash and Claude Sonnet 4.6 exhibited similar improvement trends. These findings indicate that evaluation rubrics, clinical knowledge, penalty rules, and structured outputs contribute to improving the clarity, consistency, and reproducibility of LLM-based visual explanation evaluation.

Nevertheless, this study has several limitations. First, it was conducted as a pilot study using only a single representative facial atopic dermatitis image. Second, both the number of evaluation samples and the number of human evaluators were limited. Therefore, the findings should be interpreted as preliminary evidence demonstrating the feasibility of applying LLM-as-a-Judge frameworks to medical image explanation evaluation.

Furthermore, the current study did not establish an expert-derived ground truth for explanation quality. Consequently, direct validation against dermatologist assessments was not possible. Future work will involve constructing a benchmark dataset containing more than 100 facial skin disease images covering acne, seborrheic dermatitis, and atopic dermatitis. Large-scale Visual Attention Map explanation samples generated from three CNN architectures and four augmentation strategies will be evaluated. In addition, expert dermatologists and medical AI researchers will be recruited to establish reliable ground-truth annotations, enabling more rigorous validation of LLM-based explanation evaluation frameworks.